\documentclass[conference]{IEEEtran}

\usepackage{graphics} 
\usepackage{epsfig} 
\usepackage{mathptmx} 
\usepackage{times} 
\usepackage{amsmath} 
\usepackage{amssymb}  
\usepackage{array} 
\usepackage{booktabs} 
\usepackage{colortbl} 
\usepackage{multirow}
\usepackage{xcolor}
\usepackage[linesnumbered,algoruled,boxed,lined]{algorithm2e}
\usepackage{textcomp}
\usepackage{graphicx}
\usepackage{subcaption}

\def\BibTeX{{\rm B\kern-.05em{\sc i\kern-.025em b}\kern-.08em
    T\kern-.1667em\lower.7ex\hbox{E}\kern-.125emX}}
    
\usepackage[colorlinks=true,citecolor=blue,bookmarks=false]{hyperref}
\usepackage[capitalise]{cleveref} 

\newcommand\blfootnote[1]{%
  \begingroup
  \renewcommand\thefootnote{}\footnote{#1}%
  \addtocounter{footnote}{-1}%
  \endgroup
}

\begin{document}

\title{\LARGE \bf
Table-Of-Contents generation on contemporary documents}

\author{\IEEEauthorblockN{Najah-Imane Bentabet\textsuperscript{*}}
\IEEEauthorblockA{\textit{Fortia Financial Solutions} \\
17 Av George V, Paris, France \\
najah-imane.bentabet@fortia.fr}
\and
\IEEEauthorblockN{R\'emi Juge\textsuperscript{*}}
\IEEEauthorblockA{\textit{Fortia Financial Solutions} \\
17 Av George V, Paris, France \\
remi.juge@fortia.fr}
\and
\IEEEauthorblockN{Sira Ferradans}
\IEEEauthorblockA{\textit{Fortia Financial Solutions} \\
17 Av George V, Paris, France \\
sira.ferradans@fortia.fr}
}

\maketitle
\thispagestyle{empty}
\pagestyle{empty}

\begin{abstract}

The generation of precise and detailed Table-Of-Contents (TOC) from a document is a problem of major importance for document understanding and information extraction. Despite its importance, it is still a challenging task, especially for non-standardized documents with rich layout information such as commercial documents. 
In this paper, we present a new neural-based pipeline for TOC generation applicable to any searchable document. Unlike previous methods, we do not use semantic labeling nor assume the presence of parsable TOC pages in the document. 
Moreover, we analyze the influence of using external knowledge encoded as a template. We empirically show that this approach is only useful in a very low resource environment.
Finally, we propose a new domain-specific data set that sheds some light on the difficulties of TOC generation in real-world documents. 
The proposed method shows better performance than the state-of-the-art on a public data set and on the newly released data set.

\end{abstract}

\section{INTRODUCTION}
\blfootnote{\textsuperscript{*}Both authors contributed equally to this work}
Long document comprehension is still an open problem in Natural Language Processing (NLP). 
Most of the corporate information or academic knowledge is locked in long documents ($>10$ pages) with complex semantic and layout structure. Documents are generally converted into plain text and processed sentence by sentence, where the only structure that is easily identified are the paragraphs, thus loosing the internal organization of the document. But the common hierarchical structure known as Table of Content (TOC) is of fundamental importance for NLP tasks related to discourse structure analysis or topical extraction, such as Information Extraction or Question Answering. It also improves the reader’s user experience, since it allows to better navigate the document

In this paper we focus on the generation of the document's TOC out of its content relying on layout features and text. 

The TOC of a document is the hierarchical structure of its titles, that is, the tree structure that defines the hierarchical dependencies between the sections. We call the nodes of this tree, logical entities. 
We find two ways of labeling the TOC: logical labels and semantic labels. A logical label assigns to a title the depth position in the TOC tree. 
On the other hand, semantic labeling references to a type of logical entity. 
Within a single type of document, we can find logical entities that have the same semantic role, for instance, \emph{Numerical results} or \emph{Experimental results} in a scientific paper. Tagging these logical entities with the same label is what is referred to as semantic labeling of the logical entity\footnote{In the literature the difference between logic label and semantic label is not clear, see for instance~\cite{paass2011machine,constantin2013pdfx}. As we will point out along the paper, this is an important distinction.}. It is a common practice to use the semantic labels (or \textit{a priori} information) to improve the estimated tree structure using a predefined TOC (i.e. \emph{Abstract} is a section of the document below \emph{Title}), see for instance~\cite{paass2011machine,constantin2013pdfx}. 
Unlike these methods, we disentangle logical and semantic label estimation. This allows us to avoid the usage of this rigid semantic structure of the document, and thus, tackle different types of documents with the same method. Instead we just set a maximum depth of the logical tree structure.


\textbf{Previous work.} Previous work explored TOC generation from mainly three perspectives. It was first approached from a \emph{layout analysis} point of view: a document image is decomposed into a variety of physical predefined entities such as text, figures, tables, and background~\cite{namboodiri2007document}. Each entity is characterized by a set of predefined features~\cite{conway1993page,fourli2007bayesian,nakagawa2004extraction} and a semantic predefined label (titles, captions, author names,...). The semantic labels are assigned using heuristic rules~\cite{conway1993page} or classification methods~\cite{tsujimoto1990understanding}. The given hierarchical structure between these entities provides the TOC~\cite{paass2011machine}. 
These approaches are highly constrained since they need a predefined set of semantic elements, normally very connected to a type of document, and a reference TOC which is not always available or applicable. These methods were mostly applied to scientific papers.

More recent algorithms focus on automatic TOC \emph{extraction} where the goal is to parse this hierarchical structure of sections and subsections from the TOC pages embedded in the document. Most of the research developed in this area has been linked to the INEX~\cite{Dresevic2009} and ICDAR competitions~\cite{doucet2013icdar,beckers09,nguyen2018enhancing} which target old and long OCR-ised books instead of small papers as for the previous methods. Outside  these competitions, we find the methods proposed by Elhaj et al~\cite{elhaj2014,elhaj:2018}, based also in TOC page parsing.

Finally, we find methods that detect headers using learning methods based on layout and text features. The set of headers are hierarchically ordered according to a predefined rule-based function~\cite{doucet2013icdar,liu2011toc,gopinath2018supervised}.

\textbf{Limitations of previous works.}
The methods in the state of the art require either a highly detailed semantic reference TOC, or the presence of a TOC page in the input document.

The first group of methods (~\cite{namboodiri2007document,conway1993page,fourli2007bayesian,nakagawa2004extraction}) highly restricts the usage of the algorithm since it can only target one type of document at a time (for instance scientific papers in \textit{Arxiv}).
Unlike these methods, we disentangle logical and semantic label estimation. This allows us to avoid the usage of this rigid hierarchical semantic structure of the document, and thus, deal with different types of documents with the same method. Instead we assume a maximum depth of the logical tree structure. 

The second type gives for granted a TOC page~\cite{doucet2013icdar,beckers09}, which may not appear in the document. We do not assume either the presence of a TOC page that can be parsed. Instead, we detect and order the titles in the document. The main advantage of this approach is that the tree depth is limited by the titles that appear in the document, and not by the tree displayed in the TOC page, which tends to be narrower. 


We propose a method that learns how to sequentially order the titles of a document into a tree structure without any reference TOC. To the best of our knowledge, this approach has not been explored in the literature before.
The contributions of the paper can be summarized as follows: 
\begin{itemize}
 \item We present a new model for multi-scale document TOC generation that can handle documents with flat and deep hierarchy, spanning small or large number of pages, and using both the title and the document level.
 \item We show the performance of the proposed TOC generation method on a public data base as well as on a new domain-specific data set.
 \item We lead a thorough analysis of the integration of external information provided by an external template in a low resource domain.
\end{itemize}

\section{PROPOSED METHOD}\label{sec:pipeline}

We present a TOC-generation pipeline comprising two building blocks: the first unit is a binary classifier that distinguishes titles from non-titles, while the second unit is made of a sequence labeling model that orders hierarchically the previously detected titles. 

In the following sub-section, we describe how we represent the input of our pipeline. Then, we explicit the hand-crafted features used to better represent the input before detailing the models used in both units in subsections ~\ref{ssec:charcnn_title_detector} and \ref{ssec:title_hierarchization} respectively. Finally, we explain the last step of the pipeline that produces the final TOC. Algorithm \ref{algo:TOC_gen} details the complete algorithm that is presented in this paper. 

\begin{algorithm}[!htb]
\SetAlgoLined
 \SetKwInOut{Input}{Input}
    \SetKwInOut{Output}{Output}
    \Input{$D$: Input document}
    \Output{$\{(s_i,h_i)\}_{i=0}^N$: Predicted TOC where:
\begin{itemize}
    \item[] $s_i$ is the $i$-th detected title (string)
    \item[] $h_i$ is the hierarchical level predicted for $s_i$
\end{itemize}}
  
$S = \emptyset$ \tcp{List of titles}
$T$ := $\{(t_j,l_j)\}_{i=0}^M$ = segmentation($D$) \tcp{see sect.\ref{ssec:input_representation}} 
\For{$(t_j,l_j)$ in $T$}{
    $f_j$ = extract\_hand\_crafted\_features($t_j,l_j$) \tcp{see sect.\ref{ssec:hfs}}
    \If{is\_title($t_j$,$f_j$)}{\tcp{see sect.\ref{ssec:charcnn_title_detector}}
        add\_to\_title\_list($t_j$,$f_j$,$S$)
    }
}
\For{$s_i$ in $S$}{
    $\hat h_i$ = get\_hierarchy($s_i$, $S$)  \tcp{see sect.\ref{ssec:title_hierarchization}}
}
$\{(s_i, h_i)\}_{i=0}^N$ = toc\_postprocessing($\{(s_i, \hat h_i)\}_{i=0}^N$)  \tcp{ see sect. \ref{ssec:toc_gen}}
\caption{TOC-generation algorithm}
\label{algo:TOC_gen}
\end{algorithm}

\subsection{Document preprocessing and segmentation} \label{ssec:input_representation}
The document content is segmented into a set of visually similar-looking regions, that we call \textit{text blocks}, using a rule-based method.
Apart from \cite{Ramakrishnan2012,tuarob2015,Budhiraja2018}, most segmentation literature focuses on document images (e.g \cite{Zahour2009,barlas:hal-00981245})

In this paper, we choose to use a simple layout-based segmentation method. 
The first step is to discard images, footers and headers using predefined heuristics. More specifically, we discard all text that is not in the center of the page, defined by a manually set of thresholds.

Once we have the page content cleaned from headers and footers, we proceed to group together consecutive and similar-looking text lines which constitute the set of text blocks. Unlike \cite{Budhiraja2018}, we choose to classify regrouped text blocks instead of text lines because our data set titles may span multiple lines. 

\begin{figure}[t!]
    \centering
    \includegraphics[scale=.6]{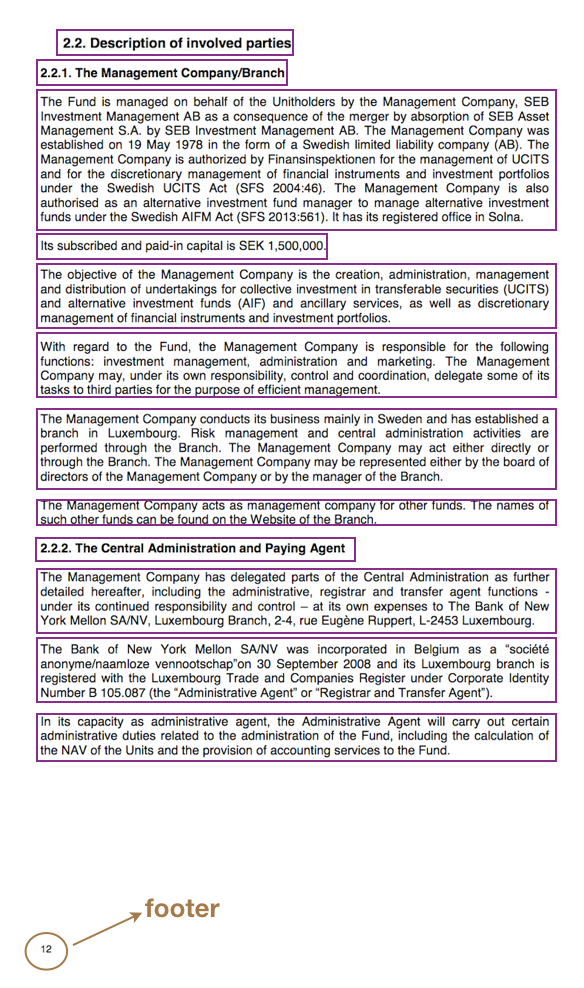}
    \caption{Output of the segmentation method on a sample page document; purple rectangles delimits the text blocks computed by our segmentation method.}
    \label{fig:segmentation}
\end{figure}

The entries in the generated TOC are text blocks. \cref{fig:segmentation} shows a typical output of the segmentation method.

\subsection{Hand-crafted features}\label{ssec:hfs} 
Each text block is represented with a set of 28 hand-crafted features listed in \cref{tab:hfs}. 

\begin{table}[h!]
\caption{Hand-crafted features used to represent text blocks}
\begin{tabular}{|p{4cm}|p{4cm}|}
\hline Unary features & Relative features \\
\hline contains\_verb, is\_bold, is\_italic, is\_all\_caps, \textbf{text\_length}, begins\_with\_numbering, 
one\_hot, \textbf{indent}, \textbf{font\_size}  & \textit{style\_to\_prev}, \textit{style\_to\_subs}, \textit{weight\_diff\_to\_prev}, \textit{weight\_diff\_to\_subs}, \textit{size\_to\_prev}, \textit{size\_to\_subs}, \textbf{size\_diff\_to\_prev}, \textit{size\_diff\_to\_subs}, \textit{indent\_to\_prev}, \textit{indent\_to\_subs}, \textbf{indent\_diff\_to\_prev}, \textit{indent\_diff\_to\_subs}, \textbf{dist\_to\_prev\_line}, \textit{dist\_to\_subs\_line}, prev\_tb\_one\_hot, subs\_tb\_one\_hot, color\_diff\_to\_prev, color\_diff\_to\_subs   \\
\hline
\end{tabular}
\label{tab:hfs}
\end{table}

We find three different types: 
\begin{itemize}
    \item layout features: They represent visual information related to the layout (i.e. $is\_bold, is\_italic$...)
    \item semantic features: They are used to enrich the semantic representation of the titles (i.e. $one\_hot, prev\_tb\_one\_hot$...). They are all 100-dimension one-hot vectors encoding the presence of the most common tokens found in the titles of the training set.
    \item text-related features: They encode the information which is neither visual, nor semantic (i.e. $text\_length, contains\_verb$...).
\end{itemize}

These features can be unary, which means that they describe characteristics of text blocks themselves, or relative, that is, they compare characteristics of text blocks with their neighbors.

The features highlighted in bold in \cref{tab:hfs} are continuous or discrete with infinite support. These are fuzzified into three categories (small, normal, and large) using the mean and standard deviation over each document. The ones highlighted in italic in \cref{tab:hfs} have finite support. They are one-hot encoded for later use. All hand-crafted features are further explained in \cref{tab:hfs_explanation} of the Appendix.

These features ($f_j$ in Algorithm \ref{algo:TOC_gen}) are used to enrich the representation of the text blocks used by the title detection and sequence labeling steps described below. 

\subsection{Title detection}\label{ssec:charcnn_title_detector}

The first unit of our pipeline detects titles from plain text by binary classification of each text block. We use a character-level convolutional neural network (char-CNN) which has proven to be effective in text classification tasks~\cite{zhang2015char}. This choice is further motivated by the fact that titles, unlike plain text, generally begin with specific numbering schemes (e.g "I." in "I. This is a title", and "II.2.a" in "II.2.a This is a sub-title"), which is a relevant clue to distinguish titles from non-titles. 


\textbf{Title encoding.} The text blocks ($t_j$ in Algorithm~\ref{algo:TOC_gen}) are pre-processed such that punctuation is standardized. Then, the characters are encoded through an embedding layer into a $d_c$-dimensional dense vector such that each text block is represented by a $d_c \times l_c$ matrix ($l_c$ being the maximum number of characters in the text blocks).

\textbf{Classifier.} The char-CNN applies, in parallel, $n\_conv_c$ 1D-convolutions followed by 1D-max-pooling. The outputs of these parallel operations are concatenated and flattened before applying a fully-connected layer with ReLU activation. We append to the resulting vector a fixed-length vector of hand-crafted features (cf \cref{ssec:hfs}) describing the layout and the text of the text block. Finally, a second fully-connected layer with softmax activation classifies the input text block into title or non-title. We use dropout~\cite{dropout} for regularization purposes. 


\subsection{Title hierarchization}
\label{ssec:title_hierarchization}
Once the titles are detected, the role of the second unit is to hierarchically order them (i.e predict a level for each one of them such that the sequence of numbers represents a TOC tree) to create the final TOC. We model this as a sequence labeling problem, where each class corresponds to a level. 

\textbf{Title encoding.} 
Each title is encoded using an architecture similar to the one aforementioned in \cref{ssec:charcnn_title_detector}
, except that it works at the word level
(word-CNN)~\cite{kim2014cnn}. 
We use $d_w$ to denote the dimension of word embeddings. This encoder allows us to induce a low and dense vector representation of each detected title of the document, to which is appended a fixed-length vector of hand-crafted features (cf \cref{ssec:hfs}).
The resulting vectors are stacked to create a matrix ($M$) that represents the document where each row corresponds to the vector representing a title. Therefore, two consecutive titles in the document have their representations located as consecutive rows in the matrix $M$.

\textbf{Sequence labeling.} Then, we sequentially label the titles by applying a BiLSTM~\cite{LSTM_} layer followed by a CRF~\cite{Lafferty2001_crf} layer row-wisely on $M$. The first one exploits the information of both past and future input features in the document. The second one uses the level of the other titles in the document to make the predictions.

To the best of our knowledge, this is the first time title levels are predicted using a sequential labeling approach. 
We emphasize that to this model, a data point is not a single text block but a complete document, represented by the sequence of its text blocks ($D$ in Algorithm~\ref{algo:TOC_gen}). 
%
\subsection{Reorganization into a TOC}\label{ssec:toc_gen}
The sequence $\{S=(s_i, \hat h_i)\}_{i=0}^N$ is not necessarily a tree, therefore the last step of the TOC-pipeline is a post-processing step that maps the sequence $S$ into a TOC $\{(s_i, h_i)\}_{i=0}^N$ that verifies the constraints of a tree graph.
This mapping is a rule-based deterministic function. The first detected title is always rearranged as a top title whatever the level predicted for it by the BiLSTM-CRF. Then, for each subsequent detected title $s$, we look for the closest previous title $s'$ that has a predicted level (strictly) lower than that of $s$, then we set $s'$ to be the parent of $s$.
\section{DATASETS}

In our experiments, we considered two data sets (a) a small, domain-specific database of investment documents built specifically for this work, and (b) a public data set used previously to evaluate a TOC extractor on \textit{Arxiv} scientific publications \cite{rahman2017}. 

\subsection{Investment documents data set}
Investment documents are PDF documents where investment funds detail their legal structure and investment offers. These documents are good examples of domain-specific commercial documents, with specific vocabulary
and ellipsis of relevant information assumed by the context. 

\begin{figure}[b!]
  \begin{subfigure}{\linewidth}
  \centering
  \includegraphics[width=.45\linewidth]{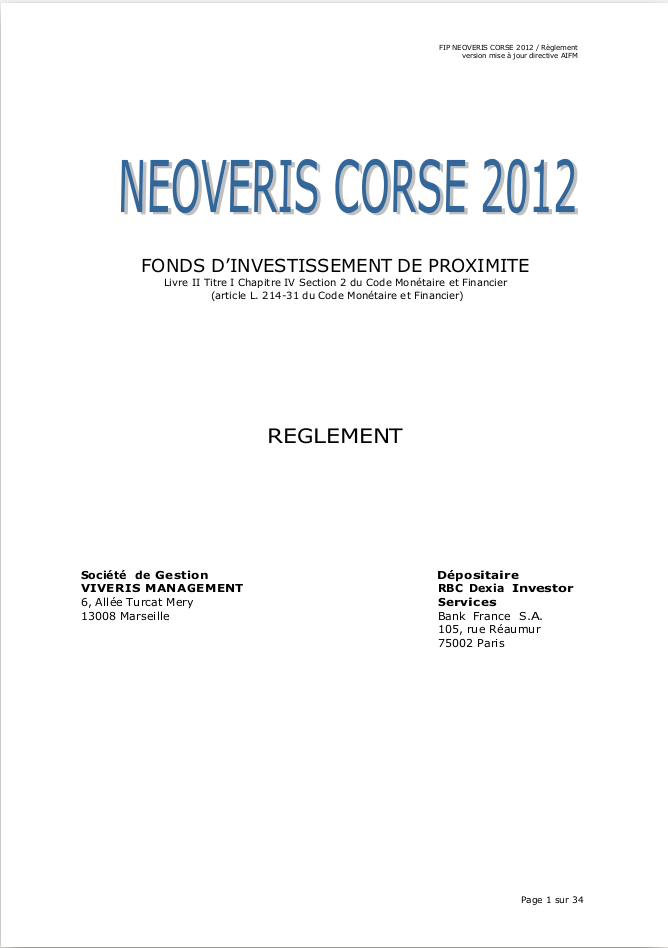}
  \includegraphics[width=.45\linewidth]{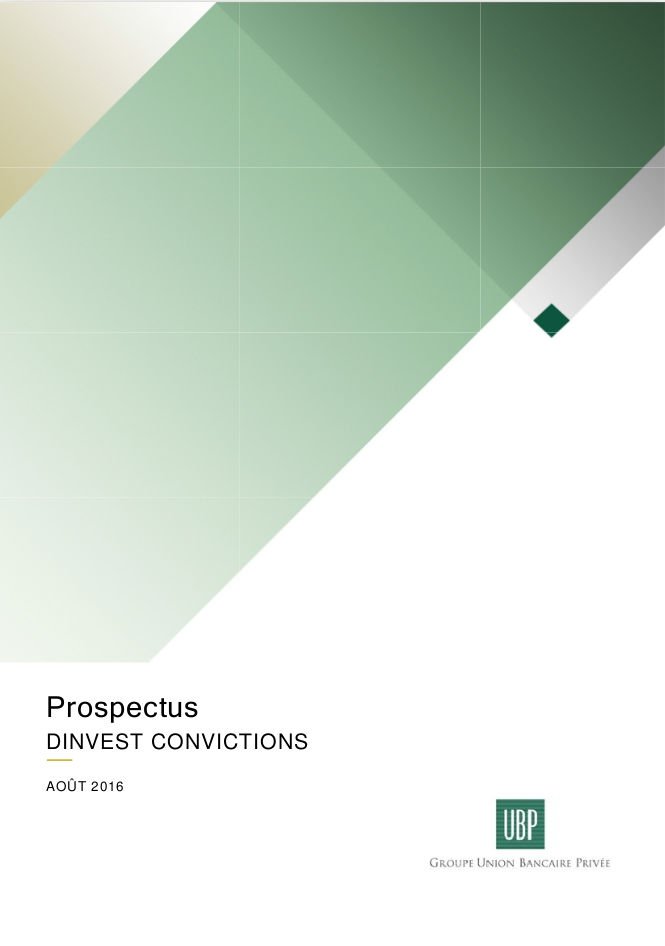}
  \end{subfigure}\par
  \begin{subfigure}{\linewidth}
  \centering
  \includegraphics[width=.45\linewidth]{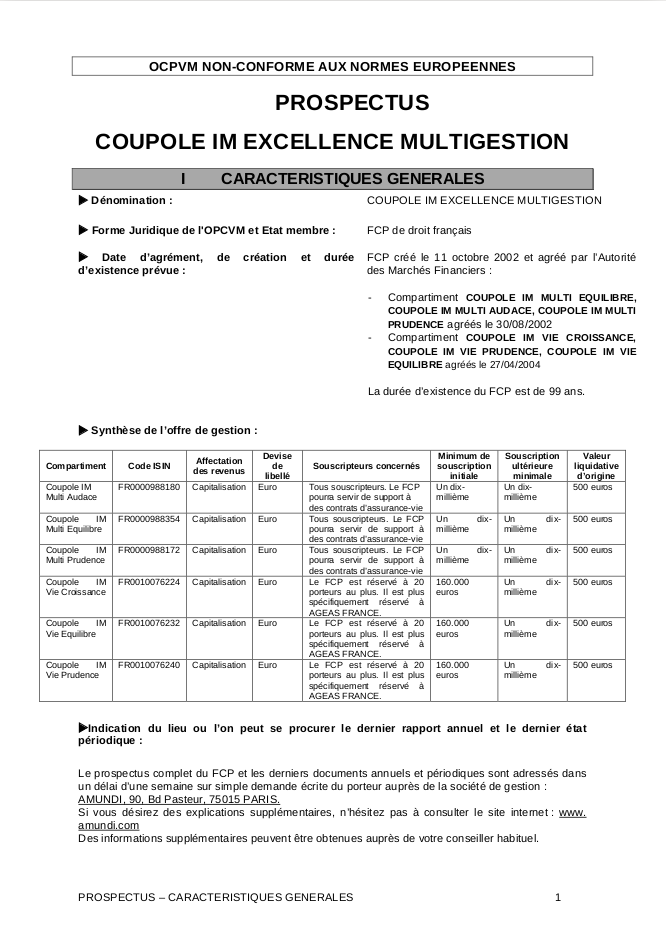}
  \includegraphics[width=.45\linewidth]{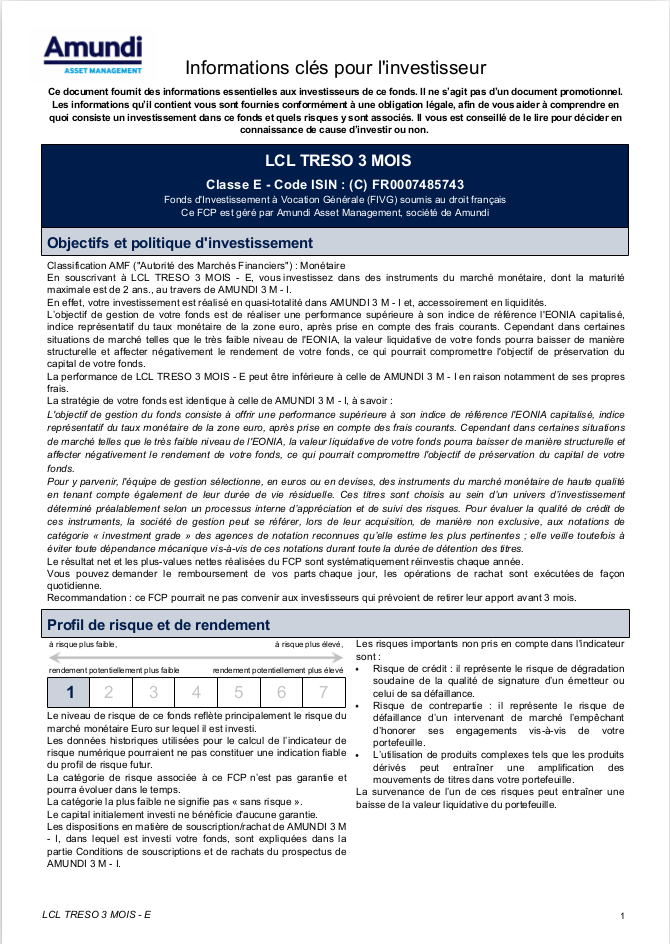}
  \end{subfigure}
  \caption{Random pages from the investment document dataset.}
  \label{fig:sample_pages}
\end{figure}

The format in which investment documents is displayed changes from one publisher to another. We observe great differences in font size, color, and style, usage of page headers and footers, as well as tables of various forms. In \cref{fig:sample_pages}, one can qualitatively assess the variety in design of french investment documents.


The french investment documents we consider in this data set are characterized by a deep hierarchy in their TOC (up to 6 levels) and are not embedded with parsable TOC pages. Titles generally contain the same words but are written differently from one document to another. Some documents contain specific titles that do not exist in other documents. Moreover, similar titles in two different documents may have different hierarchical levels. The average number of pages per document in this collection is 26 pages.

\textbf{Tagging protocol}. We developed internally a tool to manually annotate the TOC of any document in the PDF format. The tool produces a hierarchical \textit{json} file containing for each toc-entry of the TOC: its title, start-page, end-page, and children.

Each investment document was independently annotated by two people, then a third person reviewed both annotations and validated one final TOC per document, which is then used as ground truth for training and evaluating our models. The annotations were restricted such that the maximum depth of the TOC trees was set to 5. We annotated in total 71 french investment documents. Following \cite{doucet:hal-01070398}, the agreement scores between each annotator is given in terms of Inex08 F1. We also add the Xerox F1 \cite{dejean2010} scores (also referred to as 'hyperlink' score in the literature) in \cref{tab:agreement_score}. 
These scores were used in the ICDAR 2013 Book Structure Extraction competition\cite{doucet2013icdar} to compare two versions of the TOC of the same document. We can observe high agreement scores, allowing us to be confident enough about the quality of our dataset.
\begin{table}[t!]
\caption{Agreement scores between different annotators of the investment document dataset}
\centering
\begin{tabular}{|c|c|c|}
\hline & Xerox F1 & Inex08 F1 \\
\hline tagger 1 \& tagger 2 & 89.8\% & 77.0\% \\
\hline tagger 1 \& reviewer & 92.1\% & 82.8\% \\
\hline tagger 2 \& reviewer & 90.1\% & 79.6\% \\
\hline
\end{tabular}
\label{tab:agreement_score}
\end{table}

\subsection{\textit{Arxiv} dataset}
In 2017, Rahman and Finin~\cite{rahman2017} published a data set of scientific papers from \textit{Arxiv} along with their TOC annotations. It gathers all the scientific papers uploaded on the \textit{Arxiv} website between 2010 and 2016. Unlike the previous data set, only few titles are repeated in all documents (i.e."Introduction", "References", and "Conclusions"). The remaining titles are specific to the topic of the publications. We also observed that for this data set, the TOC trees are usually not as deep as in the previous data set.


\section{IMPLEMENTATION DETAILS}
\subsection{Tools}
The data sets we use are composed of searchable PDF documents. We segment them to text blocks after converting them to \emph{XML} files with \emph{pdftohtml} utility provided by the Poppler library\footnote{\href{https://poppler.freedesktop.org/}{poppler.freedesktop.org}}. We put emphasis on the fact that our method is not restricted to PDF documents, and can be used with any documents for which we can extract some meaningful layout and textual features. We mainly use the Python library Keras \cite{chollet2015keras} and the XGBoost package \cite{Chen2016_xgboost} to implement the baselines and models of this paper. 

\subsection{Hyper-parameter settings}\label{ssec:grid_search}
As explained in \cref{sec:pipeline}, our TOC generator is composed of two independent and trainable units. We optimized the hyper-parameters of each using a grid-search strategy. 
For the title detector (resp. the sequence labeling model), the best performing parameters can be observed in \cref{tab:params_title_detection} (resp. \cref{tab:params_title_classifier}). 
Cells detailing the parallel convolutions parameters use the following nomenclature: ($number\_of\_kernels, filter\_size, pool\_size$).
For both units, we used the Adam optimizer \cite{diederik2014} with a categorical cross-entropy loss. 

\begin{table}[t!]
\caption{Hyper-parameters setting of our CharCNN-based title detector.}
\begin{tabular}{|p{3.3cm}|p{2.2cm}|p{2.2cm}|}
\hline \textbf{Hyper-parameters} & \textbf{Investment document dataset} & \textbf{Arxiv dataset} \\
\hline sequence of characters length $l_c$ & $50$ & $100$ \\
\hline character embedding size $d_c$ & $200$ & $200$ \\
\hline character embedding initialization & random orthogonal & random orthogonal \\
\hline parallel convolutions ($n\_conv_c=2$) & $(16, 2, 2), (32, 5, 2)$ & $(16,2,2),(32,5,2)$ \\ 
\hline fully-connected layer output dimension & $256$ & $256$ \\
\hline dropout probability & $0.25$ & $0.0001$ \\
\hline
\end{tabular}
\label{tab:params_title_detection}
\end{table}

\begin{table}[t!]
\caption{Hyper-parameters setting of our sequence labeling model.}
\begin{tabular}{|p{3.3cm}|p{2.2cm}|p{2.2cm}|}
\hline \textbf{Hyper-parameters} & \textbf{Investment document dataset} & \textbf{Arxiv dataset} \\
\toprule
\multicolumn{3}{|c|}{Word-CNN based text encoder} \\
\toprule
sequence of words length $l_w$ & $70$  & $30$\\
\hline word embedding size $d_w$ & $300$ & $300$\\
\hline word embedding initialization & random orthogonal & random orthogonal\\
\hline parallel convolutions ($n\_conv_w=2$) & $(16, 3, 2),(16, 5, 2)$ & $(16, 3, 2),(16, 5, 2)$\\
\hline fully-connected layer output dimension & $512$ & $512$\\
\hline dropout probability & $0.25$ & $0.0001$\\
\toprule
\multicolumn{3}{|c|}{BiLSTM-CRF} \\
\toprule
number of units & $70$ & $70$\\
\hline recurrent dropout & $0.1$ & $0.1$\\
\hline
\end{tabular}
\label{tab:params_title_classifier}
\end{table}

\section{RESULTS}\label{sec:results}

\subsection{Evaluation measures}
In order to evaluate each unit alone, we used the weighted F1 score \cite{scikit-learn}. The weighted F1 score looks at the correct rate of classification at the title-level, without taking into account the hierarchy induced by the TOC. 
To address that problem, we then score in \cref{ssec:pipe_Xrx} the whole pipeline with the Xerox F1 score and the Xerox title accuracy, which allow us to assess the quality of the whole pipeline (from segmentation up to the generated TOC). 

\subsection{Tested models}\label{ssec:models_tested}
For each unit, we use as baselines Gradient Boosted Trees (GBT) models \cite{Chen2016_xgboost} on characters, on hand-crafted features, and on a combination of them. We also consider the models of Rahman and Finin ~\cite{rahman2017} as a stronger baseline for each block of the pipeline. For both title detection and title hierarchization, they used many-to-one recurrent neural networks (RNNs) applied to one-hot representations of the characters. 
As no code was provided, we re-implemented their models in our experiments.

In order to assess the quality of the whole pipeline, from segmentation up to the generation of the TOC, we use the Xerox measures \cite{dejean2010}. The first pipeline uses Rahman and Finin's models. The second one, ours, is composed of the proposed models: the char-CNN for the title detection and the word-CNN BiLSTM-CRF for the title hierarchization.

\subsection{Results on the \textit{Arxiv} dataset} \label{ssec:arxiv_results}
The \textit{Arxiv} data set being very large, 
we proceed as \cite{rahman2017} and give the F1 score of a single run. After filtering out the corrupted documents and the documents without an annotated TOC, we end up with a total of 251981 documents for training and 167434 for testing.

\paragraph{Title detection}
\begin{table}[t]
    \caption{Title detection results.}
    \centering
    \begin{tabular}{|c|>{\centering\arraybackslash}p{2.2cm}|>{\centering\arraybackslash}p{2cm}|>{\centering\arraybackslash}p{2.3cm}|}
        \hline
        & \textbf{Models} & \textbf{Arxiv dataset F1} & \textbf{Investment dataset mean F1 (std)}\\
        \hline
        \parbox[t]{1.5mm}{\multirow{4}{*}{\rotatebox[origin=c]{90}{\textit{baselines}}}} & GBT hfs & 88.11\% & 94.70\% ($\pm$ 0\%)\\
        & GBT char & 87.42\% & 96.31\% ($\pm$ 0\%)\\
        & GBT combined & 91.16\% & \textbf{98.36\%} ($\pm$ 0\%)\\
        & Rahman and Finin & 75.76\% & 95.10\% ($\pm$ 0.33\%)\\
        \hline
        \parbox[t]{1.5mm}{\multirow{3}{*}{\rotatebox[origin=c]{90}{\textit{ours}}}} & & &\\
        & char-CNN & \textbf{93.80\%} & 97.60\% ($\pm$ 0.06\%)\\ 
        & & &\\
        \hline
    \end{tabular}
    \label{tab:det}
\end{table}

As shown in \cref{tab:det}, by leveraging characters and hand-crafted features, the proposed char-CNN title detector outperforms all baselines. The best implemented baseline, "GBT combined", uses the same features but has less learning capacity to encode structural information, thus yielding a lower performance than our model. 

\paragraph{Title hierarchization}
The title hierarchization results, presented in \cref{tab:clf}, shows that our algorithm outperforms the other models. We observe that the results obtained with our re-implementation of Rahman and Finin's model differ from the ones reported in their paper. We conjecture that the reason for this is threefold. Firstly, we use a different document segmentation algorithm.
Secondly, we did not have access to the same subset of data they used for training and testing.
Finally, in their paper, Rahman and Finin restricted their TOC tree to a depth of $3$, whereas we expended it to $5$.




\subsection{Results on the investment documents dataset}\label{ssec:results_investment_docs}
We randomly split our domain-specific data set so that we use 47 documents for training and 24 for testing. For this data set, we report a mean and a standard deviation of the weighted F1 score on 10 independent runs. Given the size of the previous data set, this process could not be done. 

\paragraph{Title detection}

Title detection results appear in \cref{tab:det}. The "GBT combined" baseline performs the best. The score of our proposed method is slightly worse (less than 1\%). We argue that this is due to a lack of training data. Indeed, in a scarce data environment, neural networks are more prone to overfitting than GBT models.

\begin{table}[t]
    \caption{Title hierarchization results.}
    \centering
    \begin{tabular}{|c|>{\centering\arraybackslash}p{2.2cm}|>{\centering\arraybackslash}p{2cm}|>{\centering\arraybackslash}p{2.3cm}|}
        \hline
        & \textbf{Model} & \textbf{Arxiv dataset F1} & \textbf{Investment dataset mean F1 (std)}\\
        \hline
        \parbox[t]{1.5mm}{\multirow{4}{*}{\rotatebox[origin=c]{90}{\textit{baselines}}}} & GBT hfs & 61.89\% & 70.62\% ($\pm$ 0\%)\\
        & GBT char & 60.56\% & 71.62\% ($\pm$ 0\%)\\
        & GBT combined & 64.21\% & 74.20\% ($\pm$ 0\%)\\
        & Rahman and Finin & 75.40\% & 64.60\% ($\pm$ 1.36\%)\\ 
        \hline
        \parbox[t]{1.5mm}{\multirow{3}{*}{\rotatebox[origin=c]{90}{\textit{our}}}}
        & & &\\
        & word-CNN + BiLSTM-CRF & \textbf{92.90\%} & \textbf{84.78\% ($\pm$ 0.47\%)}\\
        \hline
    \end{tabular}
    \label{tab:clf}
\end{table}

\paragraph{Title hierarchization}
As depicted in \cref{tab:clf}, our proposed model outperforms all variants and baselines both in terms of mean and standard deviation. We argue that this is due to the fact that, contrary to the baselines, our algorithm looks at the past and future input features via the BiLSTM layer and at the document level hierarchy information via the CRF layer.

\begin{table}[b!]
 \centering
 \caption{Evaluation of TOC extraction pipelines on the investment documents dataset}
 \begin{tabular}{|c|c|c|}
 \hline
 \textbf{Pipeline} & \textbf{Xerox F1 (avg)} & \textbf{Xerox Title accuracy} \\
 \hline
 Rahman and Finin & 71.7\% & 45.7\% \\
 \hline
  our method & \textbf{80.4\%} & \textbf{46.1\%} \\
   \hline
 \end{tabular}
 \label{tab:xrce_investments_docs}
\end{table}

\begin{table}[b!]
 \centering
 \caption{Evaluation of TOC extraction pipelines on \emph{Arxiv} dataset}
 \begin{tabular}{|c|c|c|}
 \hline
 \textbf{Pipeline} & \textbf{Xerox F1 (avg)} & \textbf{Xerox Title accuracy} \\
 \hline
  Rahman and Finin & 25.3\% & 28.8\% \\
 \hline
  our method & \textbf{32.8\%} & \textbf{36.4\%} \\
  \hline
 \end{tabular}
 \label{tab:pipe_arxiv}
\end{table}

\subsection{Pipeline scoring : Arxiv \& Investment documents data sets}  \label{ssec:pipe_Xrx}
\cref{tab:xrce_investments_docs,tab:pipe_arxiv} report the average Xerox F1 score and Xerox title accuracy score of the pipelines detailed in \cref{ssec:models_tested}. 
Our pipeline shows better performance than Rahman and Finin's pipeline (more than 7\% increase in F1 score on both data sets). It leverages the temporal context and sequentiality of titles thanks to the BiLSTM and CRF layers.





\section{INTEGRATION OF TEMPLATE-BASED EXTERNAL KNOWLEDGE}
\label{sssec:template_vector}

The usage of a template can be common in many areas of document creation. This is definitely the case for scientific papers, but also for commercial documents or medical reports~\cite{banisakher2018automatically}. As detailed in the Introduction, these templates are generally used in the literature for semantic labeling of the titles. The semantic label is then used to refine the logical labeling.

In this section, we would like to test the assumption that, for the hierarchization, semantic labeling based on a predefined template helps logical labeling, as it is commonly assumed in the literature. We focus on the investment documents data set as we have access to a much richer template, as detailed in the next section.

\subsection{Implementation}

The french investment documents we consider in this dataset generally follow a template\footnote{found \href{https://www.amf-france.org/en_US/Formulaires-et-declarations/OPCVM-et-fonds-d-investissement/OPCVM/Plan-type-du-prospectus0}{here}} given by the AMF\footnote{the French Financial Markets Regulator} but slightly diverge from it most of the time. 

We encode this template as a TOC tree and we compare the performance of three different methods: simple template matching, template usage in a neural model, and no-template neural model (presented in \cref{ssec:title_hierarchization}).

\textbf{Simple template matching.} We use a template-matching algorithm that matches each detected title in a document to a title in the template using a Levenshtein distance~\cite{levenshtein1966bcc}. When such matching happens with a distance below a predefined threshold, the title of the document is assigned the level of the template's title it was matched to. Otherwise, it is assigned a negative level. We can then score this rule-based algorithm and compare it to the proposed model. 

\textbf{Template usage in a neural model}. We integrate the template-matching algorithm to our proposed model in the following way: we add the result of the matching algorithm to the features representing the text blocks. The hierarchy-level assigned by the template-matching algorithm is one-hot encoded and concatenated to the feature vector extracted by our text encoder (see \cref{ssec:title_hierarchization}) and the hand-crafted feature vector (cf \cref{ssec:hfs}) . 
The model that predicts the final TOC can then use this template-level information: either it validates it or corrects it according to the input feature vectors.

\subsection{Results}




\begin{figure}[ht!]
    \centering
    \includegraphics[width=.5\textwidth]{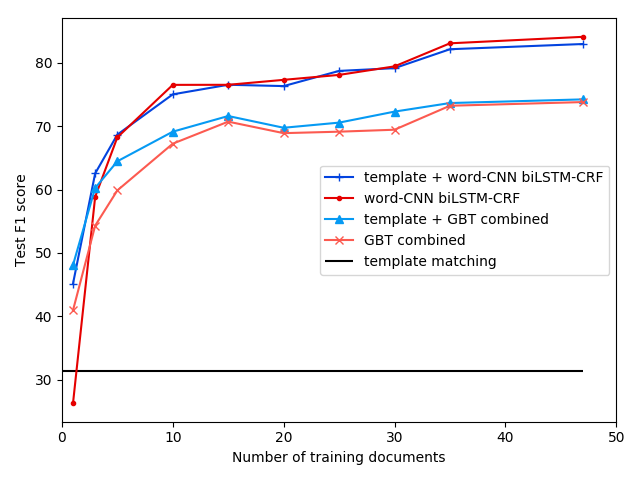}
    \caption{Learning curves obtained with and without template information.}
    \label{fig:lc_template}
\end{figure}

In Fig.~\ref{fig:lc_template}, we show the learning curves obtained with three different models (the proposed one, the GBT combined baseline and the template-matching algorithm) on the title hierarchization task with and without the template feature data. The reported error is computed on the same testing set for all five models. We can observe two regimes: low resource data set with less than 10 documents in the training set, and the higher resource regime for bigger training sets. All models that use template features (blue lines in the plot) outperform those without template features (marked as red in the figure) in a low resource regime. However, as the number of training documents augments, the choice of the model (word-CNN+bilistm-CRF vs GBT) is more important than the usage of template information.

We can, therefore conclude, that the common practice in the literature of using the semantic label to improve the TOC construction is relevant only in a very low resource regime. As the number of samples augments, this information becomes irrelevant.

\section{Conclusions}
In this paper we tackled the TOC generation problem. We proposed a neural pipeline for TOC generation that first detects the titles and then hierarchically orders them following a sequence labeling approach. The main advantage of this sequence labeling method is that the level given to a title is impacted by all the input feature vectors and the levels of the other titles in the document. The proposed method was tested on a new data set of commercial documents that are particularly challenging due to their layout complexity and TOC structure diversity. The presented algorithm outperforms largely the state of the art methods on the public \textit{Arxiv dataset} ($+18.04\%$ for the title detection and $+17.50\%$ for the title hierarchization).
Finally, we show that the common practice in the literature of using semantic labeling to improve the estimated tree structure might be unnecessary when a big data base of labeled documents is provided as a training set. 

As future work, we would like to substitute the heuristic method in our pipeline consisting of extracting raw text blocks from the documents. A relevant avenue is to replace this step with a trainable model.


\bibliography{biblio}
\bibliographystyle{IEEEtran}

\newpage
\appendix
\pagenumbering{roman}
\setcounter{page}{0}
A brief explanation of each hand-crafted feature is given in \cref{tab:hfs_explanation}
\begin{table}[h!]
\caption{Description of hand-crafted features}
\begin{tabular}{|p{4cm}|p{5cm}|}
\hline Feature & Description \\
\hline contains\_verb & does the textblock contain a verb \\
\hline is\_bold & does the textblock appear in bold \\
\hline is\_italic & is the textblock written in italic \\
\hline is\_all\_caps & is the textblock written in capital letters \\
\hline text\_length & nb of characters in the textblock \\
\hline begins\_with\_numbering & does the textblock begin with a numbering schema such as "1.", "2.a)", etc... \\
\hline one\_hot & binary vector encoding the presence/absence in textblock of most common words in training titles \\
\hline prev\_tb\_one\_hot & binary vector encoding the presence/absence in previous textblock of most common words in training titles \\
\hline subs\_tb\_one\_hot & binary vector encoding the presence/absence in subsequent textblock of most common words in training titles \\
\hline indent & horizontal distance between left margin and begining of textblock \\
\hline font\_size & size of letters \\
\hline style\_to\_prev & comparison of textblock's font style with that of the textblock immediately before it \\
\hline style\_to\_subs & comparison of textblock's font style with that of the textblock immediately after it \\
\hline weight\_diff\_to\_prev & difference between textblock's font weight and font weight of the textblock appearing immediately before it \\
\hline weight\_diff\_to\_subs & difference between textblock's font weight and font weight of the textblock appearing immediately after it \\
\hline size\_to\_prev & comparison of textblock's font size with that of the textblock immediately before it \\ 
\hline size\_to\_subs & comparison of textblock's font size with that of the textblock immediately after it \\
\hline size\_diff\_to\_prev & difference between textblock's font size and font size of the textblock immediately before it \\ 
\hline size\_diff\_to\_subs & difference between textblock's font size and font size of the textblock immediately after it \\
\hline indent\_to\_prev & comparison of textblock's indent with that of the textblock immediately before it \\ 
\hline indent\_to\_subs & comparison of textblock's indent with that of the textblock immediately after it \\
\hline indent\_diff\_to\_prev & difference between textblock's indent and indent of the textblock immediately before it \\
\hline indent\_diff\_to\_subs & difference between textblock's indent and indent of the textblock immediately after it \\
\hline dist\_to\_prev\_line & vertical distance (in pixels) between textblock and the textblock immediately before it \\ 
\hline dist\_to\_subs\_line & vertical distance (in pixels) between textblock and the textblock immediately after it \\
\hline 
\hline color\_diff\_to\_prev & 1 if color of textblock is the same that that of the textblock immediately before it, 0 otherwise \\ 
\hline color\_diff\_to\_subs & 1 if color of textblock is the same that that of the textblock immediately after it, 0 otherwise \\ \hline
\end{tabular}
\label{tab:hfs_explanation}
\end{table}

\end{document}